\documentclass[letterpaper, 10 pt, conference]{ieeeconf}  

\IEEEoverridecommandlockouts                              

\usepackage{cite}
\usepackage{hyperref}       
\usepackage{url}          
\usepackage{booktabs}       
\usepackage{amsfonts}       
\usepackage{nicefrac}       
\usepackage{microtype}      
\usepackage{xcolor, colortbl}         
\usepackage{wrapfig}
\usepackage{caption}
\usepackage{graphicx}
\usepackage{amsmath}
\usepackage{multirow}
\definecolor{lightcyan}{rgb}{0.88, 1, 1}
\definecolor{lightcoral}{rgb}{0.94, 0.5, 0.5}

\title{\LARGE \bf
AD4RL: Autonomous Driving Benchmarks for \\ Offline Reinforcement Learning with Value-based Dataset 
}

\author{Dongsu Lee$^\dagger$, Chanin Eom$^\dagger$, and Minhae Kwon$^{\dagger, *}$
\thanks{This work was supported in part by the National Research Foundation of Korea (NRF) grant (RS-2023-00278812) and in part by the ITRC (Information Technology Research Center) support program (IITP-2022-2020-0-01602) funded by the Korea government (MSIT). Dongsu Lee is grateful for financial support from Hyundai Motor Chung Mong-Koo Foundation. (Corresponding author: M. Kwon)}
\thanks{$^{\dagger}$Authors are with the Department of Intelligent Semiconductors, Soongsil University, Seoul 06978, Republic of Korea (e-mail:$\{$movementwater, eci0623$\}$@soongsil.ac.kr, minhae@ssu.ac.kr)}
\thanks{$^*$M. Kwon is with the School of Electronic Engineering, Soongsil University, Seoul 06978, Republic of Korea}
}

\begin{document}

\maketitle
\thispagestyle{empty}
\pagestyle{empty}

\begin{abstract}
Offline reinforcement learning has emerged as a promising technology by enhancing its practicality through the use of pre-collected large datasets. Despite its practical benefits, most algorithm development research in offline reinforcement learning still relies on game tasks with synthetic datasets. To address such limitations, this paper provides autonomous driving datasets and benchmarks for offline reinforcement learning research. We provide 19 datasets, including real-world human driver's datasets, and seven popular offline reinforcement learning algorithms in three realistic driving scenarios. We also provide a unified decision-making process model that can operate effectively across different scenarios, serving as a reference framework in algorithm design. Our research lays the groundwork for further collaborations in the community to explore practical aspects of existing reinforcement learning methods. Dataset and codes can be found in \hyperlink{https://sites.google.com/view/ad4rl}{https://sites.google.com/view/ad4rl}.
\end{abstract}

\section{INTRODUCTION}
Considerable progress in intelligent machines has made remarkable strides, fueled by deep neural networks trained on large datasets~\cite{goodfellow2016deep, pouyanfar2018survey, kamilaris2018deep}. In contrast, an intelligent automated system such as autonomous driving has seen limited improvement. Deep reinforcement learning was a promising solution for modern control systems~\cite{mnih2015human, sutton2018reinforcement}, but it faces challenges due to its technical characteristics. The practical challenges of online reinforcement learning are as follows~\cite{niu2022trust, levine2020offline}. Firstly, trial-and-error based learning may lead to financial loss and social damage in mission-critical systems, e.g., car accidents. Secondly, simulator-based training has an inherent gap between the simulator and real-world dynamics, resulting in limited performance in real-world deployment. Lastly, the active data collection during the training, i.e., online interaction between the agent and the environment, is expensive and hampers the ability to exploit vast previously collected datasets. Addressing these challenges is critical to realizing the full potential of reinforcement learning.

To overcome the challenges, offline reinforcement learning, also known as batch reinforcement learning~\cite{lange2012batch}, has recently gained attention as a promising approach for autonomous systems. This paradigm utilizes large-scale pre-collected datasets to train policies for agents. Since online data collection is no longer necessary for training, it allows us to avoid having agents perform immature and risky actions with an unstable policy in the early training phase. Offline reinforcement learning offers a secure and efficient learning method by leveraging insights from successful data-driven deep learning, providing the potential to shift the paradigm of mission-critical applications~\cite{fang2022offline, liang2018cirl, kumar2022workflow}.

Despite recent attention to offline reinforcement learning, research on autonomous driving still heavily relies on online reinforcement learning~\cite{kiran2021deep}. Some recent efforts have attempted to shift the research paradigm toward offline reinforcement learning, but they are still in the early stages~\cite{fang2022offline, fu2020d4rl, shi2021offline}. For example, \cite{fu2020d4rl} provides synthetic datasets and benchmarks using the FLOW framework~\cite{vinitsky2018benchmarks}. While it serves as a valuable dataset and benchmark in offline reinforcement learning studies, it focuses exclusively on acceleration maneuvers, neglecting lane-changing and presenting unrealistic, simplified driving scenarios. Additionally, it assumes that safety modules can prevent all accidents, which is an unrealistic assumption~\cite{kreidieh2018dissipating}. Another challenge with existing benchmarks is the absence of real-world datasets. Most studies rely solely on synthetic datasets collected by online reinforcement learning agents without incorporating any real-world human-driving datasets~\cite{fu2020d4rl, shi2021offline, gong2022mind}.

\textbf{Contributions:} Our primary contribution is the incorporation of real-world human-driving datasets as well as synthetic datasets in offline reinforcement learning for autonomous driving tasks. We employ the US Highway 101 dataset~\cite{USdataset} collected by the next generation simulator (NGSIM) project of the Federal Highway Administration (FHWA)~\cite{us2008ngsim}. To provide a useful dataset and benchmark, we pre-process the NGSIM dataset by labeling the reward, correcting errors, and normalizing values. We also propose a unified partially observable Markov decision process (POMDP) that can be applied across various driving scenarios. Finally, we benchmark offline reinforcement learning algorithms within the FLOW framework and extend its functionality.\footnote{
This study introduces a benchmark specifically tailored for autonomous driving, aiming to ensure widespread accessibility and reproducibility. Consistent with previous literature, the study employs a simulated environment. The decision to rely on a simulated environment is motivated by the inherent challenges associated with evaluating the performance of autonomous driving policies. Existing research suggests that off-policy evaluation approaches lack the necessary reliability~\cite{levine2020offline, fu2020d4rl, prudencio2023survey}. Consequently, policy candidates are evaluated exclusively within the simulated environment as a practical approach to mitigate the risks inherent in policy evaluation. It is worth noting that while policy evaluation takes place in the simulator, real-world datasets are incorporated for policy training.
}

\section{Related Works}
\textbf{Reinforcement Learning Based Autonomous Driving} 
The advancement of deep reinforcement learning has played a pivotal role in propelling the progress of autonomous driving research. Researchers have diligently tackled a spectrum of control tasks, encompassing acceleration, lane-changing, and intersection negotiation \cite{troullinos2021collaborative, stryszowski2020framework}. Previous studies have predominantly focused on specific roadway configurations, encompassing single-lane roads, multi-lane highways, on/off-ramps, and scenarios involving lane reduction \cite{vinitsky2018benchmarks, jang2019simulation, wu2021flow, li2022metadrive}. Nonetheless, these decision-making frameworks remain tailored to distinct scenarios, lacking universality across diverse driving contexts. The paramount objective of an autonomous driving decision-making model lies in its capacity to seamlessly and safely operate across a spectrum of driving scenarios. Consequently, this paper introduces a comprehensive decision-making process model designed to accommodate diverse driving scenarios.

 
\textbf{Reinforcement Learning Using Pre-collected Data}
There is a growing interest in leveraging pre-collected dataset-based training approaches for reinforcement learning, primarily driven by its practical constraints. The new paradigm encompasses offline reinforcement learning~\cite{lange2012batch}, imitative learning~\cite{fujimoto2021minimalist}, and imitation learning~\cite{osa2018algorithmic}, which leverage pre-collected datasets to build an initial policy. However, these methods suffer from two significant limitations. Firstly, in both training and deployment phases, offline reinforcement learning often suffers from extrapolation errors where the policy samples out-of-distribution actions that are not included in the training dataset~\cite{levine2020offline, prudencio2023survey}. This issue is commonly referred to as the distribution shift problem between the trajectory distribution of the trained policy and the dataset. Recent research attempts to mitigate this issue by introducing conservative or penalizing terms to align the policy's actions more closely with the training dataset, but such constraints may impose performance limitations~\cite{fujimoto2019off, kumar2019stabilizing, kumar2020conservative, kostrikovoffline}. Secondly, the majority of existing studies are confined to synthetic datasets generated by pre-trained policies within online reinforcement learning settings rather than utilizing real-world datasets~\cite{fu2020d4rl}. In response to these limitations and with the aim of providing a comprehensive benchmark for offline reinforcement learning, this paper assesses the performance of cutting-edge offline reinforcement learning algorithms using both human driver-generated datasets and synthetic datasets.


\textbf{Value-based Datasets for Autonomous Driving System}
In the realm of autonomous driving research using reinforcement learning, the predominant approach relies on image-based data for observation~\cite{mcallister2022control, pini2023safe, chiu2021probabilistic}. This approach offers the advantage of an end-to-end pipeline, where neural networks handle the entire process from perception to decision-making. However, this end-to-end methodology poses challenges in terms of interpretability, making it challenging to pinpoint the source of failures. In contrast, some studies opt for a value-based approach, employing sensor data as inputs for policy training rather than relying on images~\cite{vinitsky2018benchmarks, kheterpal2018flow, wu2021flow, fu2020d4rl}. This approach allows researchers to focus more directly on enhancing the decision-making capabilities of the policy without consideration of image processing ability. This study contributes value-based datasets designed for training driving policies across diverse driving scenarios. 

\section{Problem Formulations for Autonomous Driving Tasks}
This section introduces driving scenarios and datasets for offline reinforcement learning to train an autonomous driving policy. We consider the human driver dataset to capture the fundamental essence of offline reinforcement learning and synthetic datasets generated by online reinforcement learning agents. Subsequently, we look deeper into a unified POMDP that can work across driving scenarios.

\subsection{Driving Scenarios}
This subsection aims to extend the FLOW framework's driving scenario to a more realistic level by introducing three complex driving scenarios: 1) highway, 2) lane reduction, and 3) cut-in traffic. These scenarios are carefully designed to reflect real-world road environments more accurately.

\begin{wrapfigure}{r}{0.4\columnwidth}
\vspace{-0.2in}
  \begin{center}
  \includegraphics[width=0.4\columnwidth]{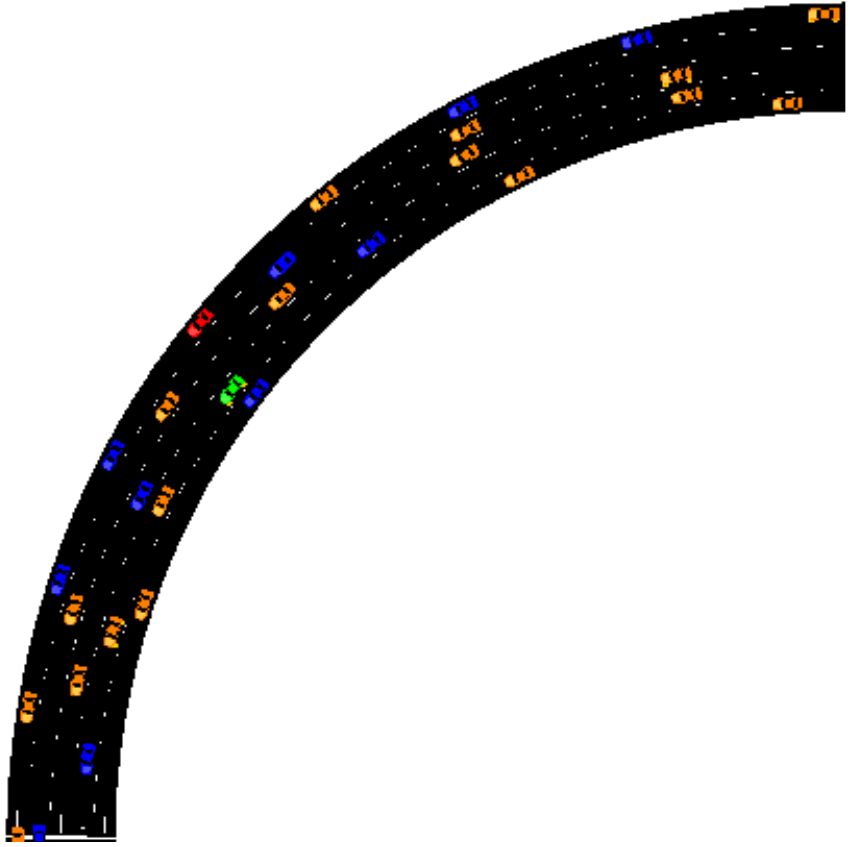}
  \end{center}
  \caption{Highway scenario}
  \label{fig:sce3}
  \vspace{-0.15in}
\end{wrapfigure}
\textbf{Highway Traffic (H):} Fig.~\ref{fig:sce3} illustrates the highway traffic scenario. 
We aim to simulate a more realistic driving environment by incorporating the patterns of the US-101 highway dataset. This includes capturing the diversified characters of vehicles on the road (e.g., desired velocity, safety distance).
The goal of the autonomous vehicle is to navigate the complex environment by making adaptive decisions, with the aim of selecting a reliable and optimal path. 

\begin{wrapfigure}{r}{0.5\columnwidth}
\vspace{-0.2in}
  \begin{center}
  \includegraphics[width=0.5\columnwidth]{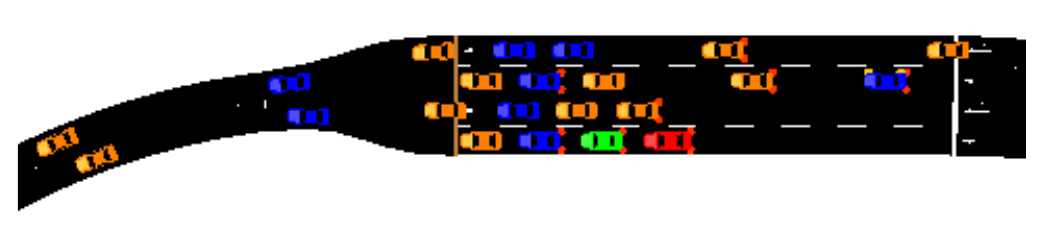}
  \end{center}
  \caption{Lane reduction scenario}
  \label{fig:sce2}
  \vspace{-0.27in}
\end{wrapfigure}
\textbf{Lane Reduction (L):}
Fig.~\ref{fig:sce2} represents the lane reduction/expansion scenario. 
Lane reduction road structures involve the conversion of a multi-lane road into a road with fewer lanes. Reducing the number of lanes limits the vehicle capacity of the road, resulting in a bottleneck that concentrates all vehicles in a narrower area and causes traffic congestion.
Therefore, drivers are forced to slow down and be more cautious. Additionally, the drivers negotiate with other vehicles to transit through the lane reduction area. The autonomous vehicle aims to find a low-density lane and move while keeping a safe distance.

\begin{wrapfigure}{r}{0.5\columnwidth}
\vspace{-0.2in}
  \begin{center}
  \includegraphics[width=0.5\columnwidth]{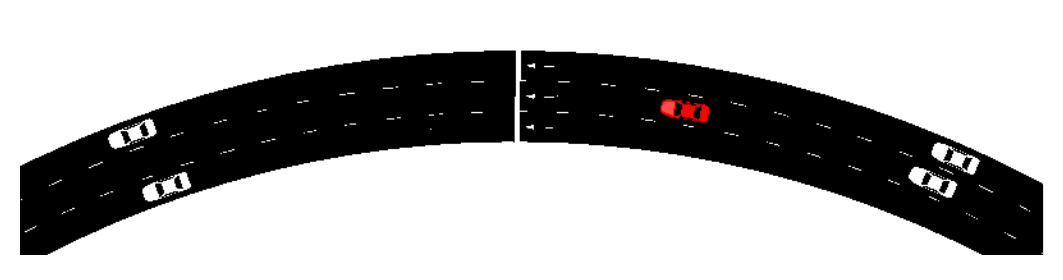}
  \end{center}
  \caption{Cut-in scenario}
  \label{fig:sce1}
  \vspace{-0.1in}
\end{wrapfigure}
\textbf{Cut-in Scenario (C):} 
In Fig.~\ref{fig:sce1}, the cut-in scenario is illustrated, which highlights the autonomous vehicle's capability to overtake other vehicles obstructing the agent's path. If the autonomous vehicle successfully overtakes other slow vehicles and maintains its desired velocity, it can maximize its rewards. To construct this scenario, we consider two setups: 1) adjusting the autonomous vehicle's desired velocity to be higher than that of the non-autonomous vehicle, and 2) regularly placing non-autonomous vehicles on the road. We expect that the autonomous vehicle will overtake other vehicles to maintain its desired velocity. 

\begin{figure}[t!]
    \centering
    \includegraphics[width=\columnwidth]{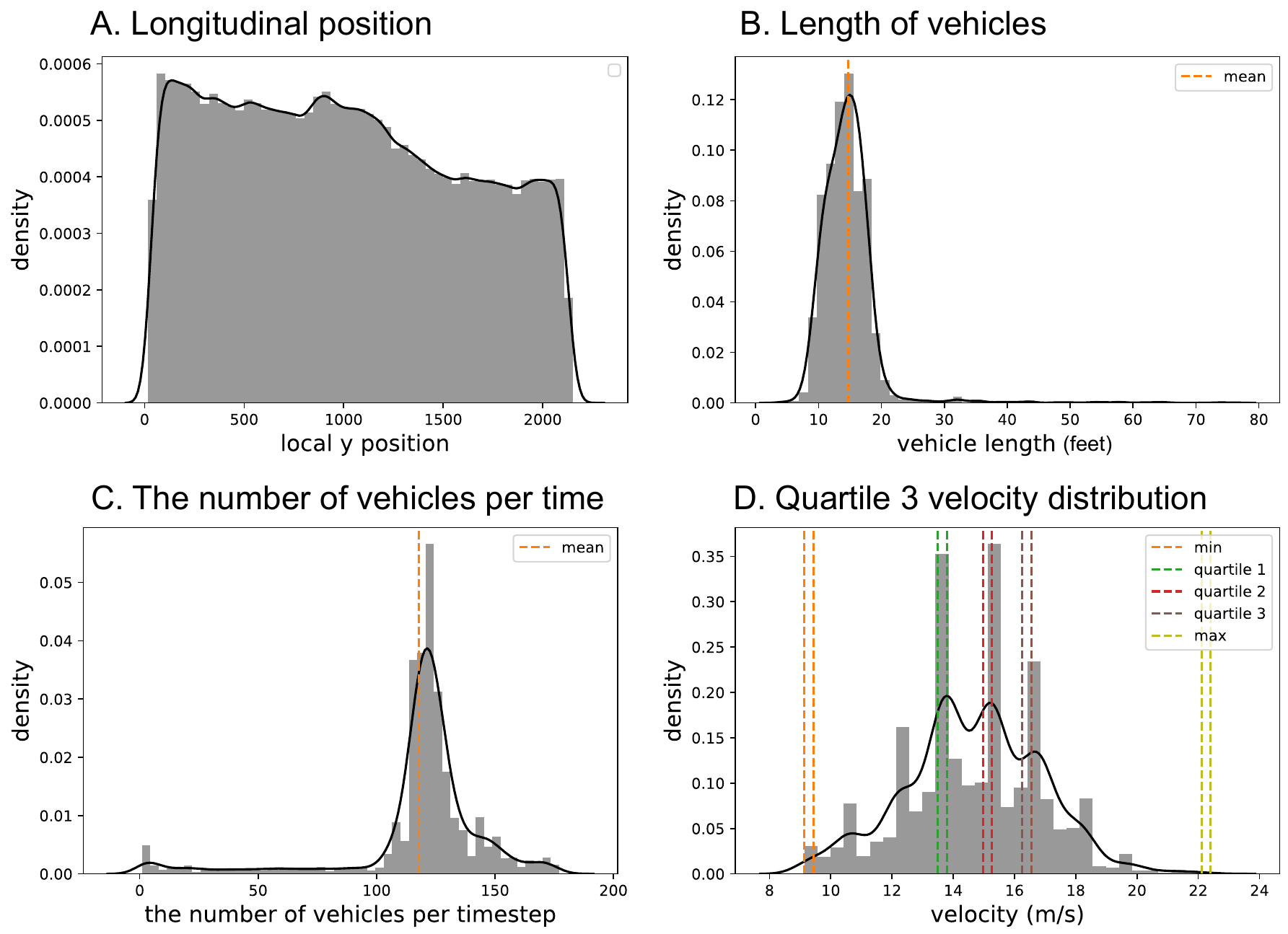}
    \caption{A: Distribution of all vehicles' longitudinal position. B: Distribution of vehicle length over all driving vehicles in dataset. C: Distribution regarding the number of vehicles per time frame. D: Distribution over quartile $3$ velocity of all vehicles. }
    \label{fig:distribution}
    \vspace{-0.5cm}
\end{figure}

To configure realistic traffic in simulation, we have analyzed the US Highway 101 dataset provided by the NGSIM project~\cite{USdataset, us2008ngsim}. We introduce several attributes and properties of this dataset (Fig.~\ref{fig:distribution}). \textbf{1) Length of road and vehicle:} Fig.~\ref{fig:distribution}\textbf{A} represents the maximum longitudinal position as approximately $2195.4$ feet, and Fig.~\ref{fig:distribution}\textbf{B} depicts the average vehicle length as approximately $14.6$ feet. \textbf{2) Target velocities:} We have first confirmed quartile $3$ of velocity distribution per vehicle. Fig.~\ref{fig:distribution}\textbf{D} shows the distribution comprising quartile $3$ velocities of all observed vehicles in the dataset. We have selected five values ($\min,~Q1,~Q2,~Q3,~\mathrm{and}~\max$) as the target velocities. \textbf{3) The number of vehicles:} Fig.~\ref{fig:distribution}\textbf{C} shows that approximately $117$ vehicles have been driving on average per time unit. Subsequently, we distribute the $117$ vehicles into five vehicle types, which mean vehicles with five different target velocities (i.e., $\min,~Q1,~Q2,~Q3,~\mathrm{and}~\max$ in Fig.~\ref{fig:distribution}\textbf{D}).

\subsection{Datasets}
\label{sec:dataset}
This subsection describes the actual and synthetic driving datasets for applying offline reinforcement learning. The number of transitions included in each dataset is approximately one million.

\textbf{Human Driver Dataset.} This research objective is to assess the feasibility and practicability of implementing offline reinforcement learning using an actual driving dataset. We utilize the Highway US-101 dataset, which is selected from the FHWA traffic analysis tool of the NGSIM Project~\cite{USdataset, us2008ngsim}.
We use the value-based dataset of NGSIM and pre-process the raw data fitted to the proposed POMDP (e.g., error correction, value normalization, and alignment with POMDP in Section~\ref{subsec:POMDP}). Note that this dataset only applies to highway traffic scenarios due to the unique properties of the roads where the data was gathered.

\textbf{Synthetic Driving Dataset.} These datasets were generated by an online reinforcement learning agent using the deep deterministic policy gradient (DDPG)~\cite{timothyddpg} algorithm, which works on continuous action and state spaces. We adopt diverse datasets to comprehend how their quality influences the performance of offline algorithms.

\begin{itemize}
    \item \textbf{Human-like}: It is a synthetic dataset generated by the Intelligent Driver Model (IDM) controller~\cite{treiber2000congested} designed to reflect the human driving pattern. This control-theoretic model focuses solely on acceleration control; to incorporate a lane-changing maneuver, we also employ the LC2013 model~\cite{erdmann2015sumo}, a manually designed lane-changing controller available within the SUMO simulator.
    
    \item \textbf{Final}, \textbf{Medium}, and \textbf{Random}: These datasets are collected by exploiting different policies, which could be obtained at different points in the training phase of online reinforcement learning. When synthesizing the ``Final" dataset, we consider the fully pre-trained policy through online reinforcement learning. Subsequently, the ``Medium" dataset is gathered using an intermediate policy obtained by an early-stopping method. The ``Random" dataset is based on the randomly initialized and unrolled policy in all scenarios.
    
    \item \textbf{Final-Medium}, and \textbf{Final-Random}: These datasets are a blend of two other datasets in equal proportions. The ``Final-Medium/Final-Random" dataset is literally combined the ``Final" and ``Medium/Random" datasets.
    
\end{itemize}

\section{Reinforcement Learning for Autonomous Driving System}

The realistic reinforcement learning problem is generally formalized as a POMDP $M=\langle \mathcal S, \mathcal O, \mathcal A,  r, \mathcal T, \Omega, \rho_0, \gamma \rangle$ that includes a state $\mathbf s \in \mathcal S$, an observation $\mathbf o \in \mathcal O$, an action $\mathbf a \in \mathcal A$, a reward $r(\mathbf s, \mathbf a,\mathbf s^\prime) \in \mathbb R$, a state transition probability $\mathcal T(\mathbf s^\prime | \mathbf s, \mathbf a)$, an observation probability $\Omega(\mathbf o|\mathbf s)$, an initial state distribution $\rho_0$, and temporal discount factor $\gamma \in [0,1)$. The agent aims to maximize the expected discounted cumulative reward $\mathbb{E}_{s_0 \sim \rho_0, \mathbf o \sim \Omega(\cdot|\mathbf s), \mathbf a \sim \pi(\cdot|\mathbf o), \mathbf s^\prime \sim \mathcal T(\cdot|\mathbf s, \mathbf a)}
\Big[\sum_t\gamma^t r(\mathbf s,\mathbf a,\mathbf s^\prime) \Big]$. 

The offline reinforcement learning samples the transitions $(\mathbf o, \mathbf a, \mathbf o^{\prime}, r)$ from the fixed dataset $\mathcal D$, thereafter minimizing an estimate of the Bellman error as follows.\footnote{In this paper, the superscript `$\prime$' implies the information on next timestep. For instance, $v_{n}^\prime$ represents the velocity of the vehicle $e_n$ at the next timestep.}
\begin{equation}
    \mathcal{L}(\theta) = 
    \mathbb E_{(\mathbf o,\mathbf a,\mathbf o^{\prime}, r) \sim \mathcal D}
    \Big[
        Q_\theta(\mathbf o,\mathbf a)-r \nonumber
        + \gamma Q_{\theta^{\prime}} (\mathbf o^{\prime}, \pi(\cdot|\mathbf o^{\prime}))
    \Big]
    \label{eq: offloss}
\end{equation}

\subsection{Unified POMDP Model}
\label{subsec:POMDP}
This subsection presents a unified POMDP structure, which can address the three different scenarios as a decision-making process for autonomous driving. We employ an actor-critic algorithm to train an autonomous vehicle as an agent on continuous observation and action spaces. The agent learns a driving policy $\pi$ that determines optimal behavior based on observable information. 
To elaborate, the agent enhances its policy by taking into account the reward signal that arises from the observation-action pair $(\mathbf{o}, \mathbf{a})$.
The primary goal of the agent is to maximize the accumulated reward.

We define road conditions as two sets: the number of vehicles and the number of lanes. The set $E = \{e_1, e_2, \cdots, e_N\}$ represents the deployed vehicles on the road. It composes the set of autonomous vehicles $E_{av}$ and the set of non-autonomous vehicles $E_{non}$, i.e., $E = E_{av} \cup E_{non}$. Subsequently, the set $K = \{1, 2, \cdots, L\}$ means the number of lanes configured on the road. 
Note that the number of lanes in specific segments can be less than the total number of lanes (e.g., lane reduction scenario).

\textbf{State}: The state $\mathbf s \in \mathcal S$ contains information about all vehicles on the road, as follows:
\begin{equation}
    \mathbf s = [v_1, p_1, k_1, v_2, p_2, k_2, \cdots, v_N, p_N, k_N]^\top, \nonumber
\end{equation}
where $v_n$, $p_n$, and $k_n$ represent the velocity, position, and lane number of vehicle $e_n$, respectively. 
When the $N$ vehicles exist on the road, the dimension of the state $\mathbf{s} \in \mathbb R^{3\times N}$.

\textbf{Observation}: The agent cannot access the complete state information, thereby relying on partial information about the state to make decisions. This constraint arises from the observability limitations of an autonomous vehicle. Specifically, the agent can observe vehicles within a restricted perceivable space $\mathcal V$ surrounding them. The perceivable space can be defined as the area that covers both the longitudinal space $\mathcal V^{long} \in [V^{long}_{\min}, V^{long}_{\max}]$ (front and behind areas) and the lateral space $\mathcal V^{lat} \in [V^{lat}_{\min}, V^{lat}_{\max}]$ (left and right sides). The vehicles in this space are defined as perceivable vehicles $e_i \in E$ of the agent $e_n$. Let us define a set of perceivable vehicles as $E_{sv}$. The set $E_{sv}$ can be formulated as $\{e_i| e_i \in E-\{e_n\}, p_i - p_n \leq |\frac{\mathcal V^{long}}{2}|, k_i - k_n \leq |\frac{\mathcal V^{lat}}{2}| \}.$

Observable vehicles are defined as the closest leading and following vehicles per visible lane among perceivable vehicles. Namely, the maximum number of observable vehicles is $2(2\mathcal V^{lat}_{\max} + 1)$, comprising $(2\mathcal V^{lat}_{\max} + 1)$ leading and $(2\mathcal V^{lat}_{\max} + 1)$ following vehicles: the set of observable leading vehicles $E_L = \{ e_{LL\mathcal V^{lat}_{max}}, \cdots, e_{LL2}, e_{LL1}, e_{LS}, e_{LR1}, e_{LR2}, \cdots, e_{LR\mathcal V^{lat}_{max}}\}$, and the set of observable following vehicles $E_F = \{ e_{FL\mathcal V^{lat}_{max}}, \cdots, e_{FL2}, e_{FL1}, e_{FS}, e_{FR1}, e_{FR2}, \cdots, e_{FR\mathcal V^{lat}_{max}}\}$

Herein, the first subscripts $L, F$ mean the leading and following vehicle; the second subscripts $L, S, R$ represent the left, same, and right lanes. To elaborate, additional subscripts of $L, R$ mean the lane gap based on the same lane $S$, i.e., $L1$ and $L\mathcal V^{lat}_{max}$ are the nearest and farthest left lane, respectively.

The observation $\mathbf o \in \mathcal O$ of the agent $e_n$ is defined as follows.
\begin{equation}
    \mathbf o = [v_n, \Delta \mathbf{v}^\top, \Delta \mathbf{p}^\top, \boldsymbol{\rho}^\top, \boldsymbol{\zeta}^\top]^\top
    \label{eq: obs}
\end{equation}
In~(\ref{eq: obs}), $\Delta \mathbf{v}$ denotes a vector of relative velocities between the agent and observable vehicles, $\Delta \mathbf{p}$ means a vector of relative distances between the agent and observable vehicles, $\boldsymbol{\rho} = [\rho_{L\mathcal V^{lat}_{max}}, \cdots, \rho_{S}, \cdots, \rho_{R\mathcal V^{lat}_{max}}]^\top$ represents a vector of lane traffic density, and $\boldsymbol{\zeta} = [\zeta_{L\mathcal V^{lat}_{max}}, \cdots, \zeta_{S}, \cdots, \zeta_{R\mathcal V^{lat}_{max}}]^\top$ depicts a vector of existence of lanes beyond the longitudinal perceivable space. 

\textbf{Action}: The vector of action $\mathbf{a} \in \mathcal A$ of the agent comprises acceleration and lane-changing, i.e., $\mathbf{a} = [a^{acc}, a^{lc}]^\top$. The acceleration behavior $a^{acc}$ is decided in continuous action space $[A_\mathrm{min}, A_\mathrm{max}]$, range between maximum deceleration and acceleration. Next, the lane-changing maneuver $a^{lc}$ is controlled in discrete action space $\{-1, 0 , 1\}$. Herein, $a^{lc} = -1$ and $a^{lc} = 1$ indicate that the agent decides to move to the right and left lane, respectively; $a^{lc} = 0$ represents keeping a lane.

\textbf{Reward}: The agent executes the action $\mathbf{a}$ in a given state $\mathbf{s}$, thereby receiving a reward $r$.  The reward is determined by a reward function, which is expressed as a linear combination of five reward components, i.e., 
\begin{equation}
 r = R(\mathbf{s}, \mathbf{a}, \mathbf{s}^\prime) = \eta_1 \mathcal{R}_1 + \eta_2 \mathcal{R}_2 + \eta_3 \mathcal{R}_3 + \eta_4 \mathcal{R}_4 + \eta_5 \mathcal{R}_5 + C. \nonumber
\end{equation}
The non-negative coefficient $\eta_n \ge 0$ refers to the weight assigned to the $n$-th reward component $\mathcal R_n$, and $C$ represents a scaling constant. 

The first reward component $\mathcal R_1$ is designed to keep the speed near the desired speed $v^*$ without overspeeding the speed limit $v_{\mathrm{limit}}$, which always satisfies $v_{\mathrm{limit}} - v^* > 0$.
\begin{align}
    \mathcal R_1
    &= 
    \begin{cases}
    \frac{v_{n}^\prime }{v^*} & v_{n}^\prime  \leq v^*\\
    \frac{v_{\mathrm{limit}} - v_{n}^\prime }{v_{\mathrm{limit}} - v^*} & v_{n}^\prime > v^* \label{eqn:R1}
    \end{cases}
    \nonumber
\end{align}
When $ 0 \leq v_{n}^\prime \leq v_{\mathrm{limit}}$, $\mathcal R_1$ becomes positive value, and if $v_{n}^\prime = v^*$, it is maximized (i.e.,  $\mathcal R_1 = 1$); when $v_{n}^\prime > v_{\mathrm{limit}}$, $\mathcal R_1$ becomes negative value as penalty. 

The second reward component $\mathcal R_2$ induces the agent to perform lane-changing that frees up driving space.
\begin{equation}
    \mathcal{R}_2 = |a^{lc}|(\Delta p_{SL}^\prime - \Delta p_{SL})
    \nonumber
\end{equation}

This component is activated when the agent performs lane-changing (e.g., $|a^{lc}| = 1$). In other words, if $|a^{lc}| = 0$, then $\mathcal{R}_2 = 0$. Subsequently, if driving space is secured after changing lanes (i.e., $\Delta p_{SL}^\prime \geq \Delta p_{SL}$), and it means desirable action, resulting in $\mathcal{R}_2 > 0$. On the other hand, if the agent fails to secure the driving space after changing lanes (i.e., $\Delta p_{SL}^\prime < \Delta p_{SL}$), thereby receiving a penalty $\mathcal{R}_2 < 0$. 

The third and fourth reward components are related to the safe distance between the leading and following vehicles, respectively. These components prevent the violation of the safe distance $s^*$.

\begin{table*}[t]
\caption{Average performance ($\pm$ confidence interval with two standard deviations) on driving scenarios and datasets. The scores are based on $10$ evaluations with five random seeds. Cyan and red highlight boxes depict the best score in each dataset and each scenario, respectively.}
\centering
\begin{tabular}{l|ccccccc}
\toprule
\scriptsize \textbf{Task Name} & \scriptsize \textbf{BC} & \scriptsize \textbf{Imitative}~\cite{fujimoto2021minimalist} & \scriptsize \textbf{BCQ}~\cite{fujimoto2019off} & \scriptsize \textbf{CQL}~\cite{kumar2020conservative} & \scriptsize \textbf{IQL}~\cite{kostrikovoffline} & \scriptsize \textbf{EDAC}~\cite{an2021uncertainty} & \scriptsize \textbf{PLAS}~\cite{zhou2021plas} \\ 
\midrule
\scriptsize \texttt{highway-US101-NGSIM} & \tiny 1202.07 $\pm$ 216.79& \tiny 1446.87 $\pm$ 242.17 & \cellcolor{lightcyan} \tiny 1501.99 $\pm$ 101.02 & \tiny 1260.59 $\pm$ 134.1 & \tiny 1261.58 $\pm$ 155.81 & \tiny 1253.35 $\pm$ 16.46 & \tiny 1468.11 $\pm$ 25.89\\ 
\scriptsize \texttt{highway-final} &\cellcolor{lightcyan} \tiny 1565.64 $\pm$ 28.03& \tiny 821.77 $\pm$ 438.46& \tiny 1563.26 $\pm$ 33.76& \tiny 1473.31 $\pm$ 162.78& \cellcolor{lightcyan} \tiny 1537.80 $\pm$ 28.13& \tiny 1219.17 $\pm$ 59.66& \tiny \cellcolor{lightcyan} 1551.31 $\pm$ 24.56\\
\scriptsize \texttt{highway-medium} & \tiny 1393.42 $\pm$ 52.39& \tiny 419.57 $\pm$ 361.79& \tiny 1402.91 $\pm$ 30.69& \tiny 1377.20 $\pm$ 73.71& \tiny \cellcolor{lightcyan} 1423.18 $\pm$ 27.39& \tiny 1244.65 $\pm$ 60.38& \tiny \cellcolor{lightcyan} 1407.75 $\pm$ 30.69\\
\scriptsize \texttt{highway-random} & \tiny 622.15 $\pm$ 234.29& \tiny -2.12 $\pm$ 13.05& \tiny 884.28 $\pm$ 93.03& \tiny -10.42 $\pm$ 14.06& \tiny 433.69 $\pm$ 49.06& \tiny 1266.66 $\pm$ 65.83& \cellcolor{lightcyan} \tiny 1543.02$\pm$ 104.65\\
\scriptsize \texttt{highway-final-medium} & \tiny 1543.33 $\pm$ 49.49& \tiny 20.87 $\pm$ 291.16& \tiny 1450.23 $\pm$ 65.49& \tiny 1434.55 $\pm$ 144.34 & \tiny \cellcolor{lightcoral} 1615.46 $\pm$ 25.18 & \tiny 1181.51 $\pm$ 53.19& \tiny 1431.30 $\pm$ 20.94\\
\scriptsize \texttt{highway-final-random} & \tiny 963.19 $\pm$ 218.79 & \tiny 25.90 $\pm$ 83.13& \tiny 659.38 $\pm$ 76.95& \tiny 240.73 $\pm$ 30.09& \tiny 835.75 $\pm$ 335 & \tiny 1261.58 $\pm$ 38.62 & \cellcolor{lightcyan} \tiny 1569.594 $\pm$ 65.28\\ 
\scriptsize \texttt{highway-human-like} & \tiny 980.80 $\pm$ 477.69 & \tiny 123.02 $\pm$ 339.94& \tiny 797.98 $\pm$ 226.53 & \tiny 766.19 $\pm$ 250.68 & \tiny 1146.09 $\pm$ 270.23 & \cellcolor{lightcyan} \tiny 1245.43 $\pm$ 4.25 & \tiny 771.11 $\pm$ 180.78 \\ 
\midrule
\scriptsize \texttt{lanereduction-final}  & \tiny 968.46 $\pm$ 36.73 & \tiny 967.75 $\pm$ 67.39 & \cellcolor{lightcyan} \tiny 1266.84 $\pm$ 240.33 & \tiny 984.36 $\pm$ 151.21& \tiny 1248.22 $\pm$ 235.20 & \tiny 268.79 $\pm$ 18.89& \tiny 965.67 $\pm$ 30.83\\
\scriptsize \texttt{lanereduction-medium} & \tiny 311.11 $\pm$ 82.90& \tiny 303.06 $\pm$ 33.04 & \tiny 1260.16 $\pm$ 280.21 & \tiny 344.17 $\pm$ 45.61& \cellcolor{lightcyan} \tiny 1319.43 $\pm$ 190.68 & \tiny 271.62 $\pm$ 18.74 & \tiny 387.01 $\pm$ 31.95 \\
\scriptsize \texttt{lanereduction-random} & \tiny 106.99$\pm$ 54.20& \tiny 139.36 $\pm$ 14.19& \tiny 31.05 $\pm$ 3.73& \tiny 3.18 $\pm$ 1.82& \tiny 312.86 $\pm$ 101.68& \tiny 269.61 $\pm$ 23.16& \cellcolor{lightcyan} \tiny 1088.64 $\pm$ 45.48\\
\scriptsize \texttt{lanereduction-final-medium} & \tiny 502.11 $\pm$ 44.96 & \tiny 790.79 $\pm$ 120.77& \tiny 1162.62 $\pm$ 276.56& \tiny 554.56 $\pm$ 83.02& \cellcolor{lightcyan} \tiny 1257.35$\pm$ 175.99 & \tiny 278.61 $\pm$ 21.85 & \tiny 366.21 $\pm$ 35.64\\
\scriptsize \texttt{lanereduction-final-random} & \tiny 157.98 $\pm$ 54.20 & \cellcolor{lightcyan} \tiny 791.42 $\pm$ 133.52& \tiny 9.82 $\pm$ 18.74& \tiny 66.40 $\pm$ 12.58& \tiny 157.27 $\pm$ 119.49& \tiny 312.99 $\pm$ 24.59& \tiny 604.6 $\pm$ 86.55\\
\scriptsize \texttt{lanereduction-human-like} & \tiny 1160.63 $\pm$ 91.30 & \tiny 960.40 $\pm$ 154.79& \cellcolor{lightcoral} \tiny 1449.62 $\pm$ 176.54 & \tiny 470.34 $\pm$ 71.13 & \cellcolor{lightcyan} \tiny 1443.84 $\pm$ 175.25 & \tiny 166.88 $\pm$ 142.23 & \tiny 1228.03 $\pm$ 8.66\\
\midrule
\scriptsize \texttt{cutin-final} & \cellcolor{lightcyan}\tiny1711.55 $\pm$ 159.27 & \tiny953.72 $\pm$ 74.21 & \tiny 1155.46 $\pm$ 215.64 & \tiny 1012.24 $\pm$ 180.62 & \cellcolor{lightcoral}\tiny 1737.22 $\pm$ 244.82 &\tiny 946.1732 $\pm$ 19.3026 & \tiny 1110.21 $\pm$ 5.94  \\
\scriptsize \texttt{cutin-medium} & \cellcolor{lightcyan}\tiny1634.74 $\pm$ 79.01 & \tiny 971.67 $\pm$ 93.78   & \cellcolor{lightcyan}\tiny 1607.19 $\pm$ 80.29 & \tiny 1201.13 $\pm$ 190.26 & \tiny 1513.23 $\pm$ 88.41& \tiny 946.1717 $\pm$ 19.3038 & \tiny 1596.30$\pm$ 24.12 \\
\scriptsize \texttt{cutin-Random} & \tiny 972.58 $\pm$ 147.16 & \tiny 1243.71 $\pm$ 257.91 & \tiny 1101.06 $\pm$ 194.16& \tiny 638.37 $\pm$ 106.86& \tiny 756.27 $\pm$ 128.54& \tiny 946.1699 $\pm$ 19.3031& \cellcolor{lightcyan}\tiny 1477.45 $\pm$ 333.78 \\
\scriptsize \texttt{cutin-final-medium} & \tiny 1084.62 $\pm$ 135.39& \tiny 866.66 $\pm$ 66.61& \tiny 1144.52 $\pm$ 196.31 & \tiny 1085.40 $\pm$ 128.06 & \cellcolor{lightcyan}\tiny 1446.98 $\pm$ 143.01 & \tiny 946.1731 $\pm$ 19.3008 & \tiny 531.35$\pm$ 41.69\\
\scriptsize \texttt{cutin-final-random} & \tiny 599.56 $\pm$ 136.57& \tiny 1135.27 $\pm$ 283.30& \tiny 1425.52 $\pm$ 310.94 & \tiny 1101.67 $\pm$ 188.81 & \tiny 798.89 $\pm$ 119.34 & \tiny 946.1719 $\pm$ 19.3014 & \cellcolor{lightcyan}\tiny 1655.24 $\pm$ 95.89  \\
\scriptsize \texttt{cutin-human-like} & \tiny 863.22 $\pm$ 166.86& \tiny 724.22 $\pm$ 113.87& \cellcolor{lightcyan}\tiny  1391.40 $\pm$ 250.34 & \tiny 1082.06 $\pm$ 178.75 & \tiny 1105.40 $\pm$ 290.50 & \tiny 946.1704 $\pm$ 19.3011 &\cellcolor{lightcyan} \tiny 1354.08 $\pm$ 19.02  \\
\bottomrule
\end{tabular}
\label{tab:flow}
\end{table*}

The third reward component is defined as follows:
\begin{equation}
    \mathcal{R}_3 = \min\left[0, 1-\left({\frac{s^*_{LS}}{\Delta{p}_{LS}^\prime}}\right)^2 \right],
    \label{eq:rew3}
\end{equation}
where $s^*_{LS}$ means that the safe distance between the agent $e_n$ and $e_{LS}$ and is defined as follows~\cite{treiber2000congested}.  
\begin{equation*} 
    s^{*}_{LS} = s_{0} + \max\left[0, v_{n}^\prime\left(t^{*}+{\frac{\Delta v_{LS}^\prime}{2\sqrt{|A_{min}\times A_{max}|}}}\right)\right]
\label{SafeDis:LS}
\end{equation*}

Herein, $s_0$ indicates the minimum safe distance between vehicles, and $t^*$ means the minimum time headway, which is the shortest time that a following vehicle can achieve without reducing velocity.
This component induces that the agent maintains the safe distance $s^{*}_{LS}$ from the leading vehicle in the same lane. Specifically, if $\Delta p_{LS}^\prime < s^*_{LS}$, then $\mathcal{R}_3$ in~(\ref{eq:rew3}) becomes the negative value; otherwise, $\mathcal{R}_3$ is not activated (i.e., $\mathcal{R}_3 = 0$). 

The fourth reward component is as follows:
\begin{equation}
    \mathcal{R}_4 = |a^{lc}|\min\left[0, 1-\left({\frac{s^*_{FS}}{\Delta{p}_{FS}^\prime}}\right)^2 \right].
    \label{eq:rew4}
\end{equation}
In~(\ref{eq:rew4}), $s^*_{FS}$ denotes that the safe distance between the agent $e_n$ and $e_{FS}$ and is defined as follows~\cite{treiber2000congested}.
\begin{equation*} 
    s^{*}_{FS} = s_{0} + \max\left[0, v_{FS}^\prime\left(t^{*}+{\frac{\Delta v_{FS}^\prime}{2\sqrt{|A_{min}\times A_{max}|}}}\right)\right]
\label{SafeDis:FS}
\end{equation*}

The same as~(\ref{eq:rew3}), $\mathcal{R}_4 \leq 0$ is always satisfied. In~(\ref{eq:rew4}), $\mathcal{R}_4$ can be non-zero only if $|a^{lc}| \neq 0$, which indicates that the agent changes the lanes. In contrast to $\mathcal{R}_3$, $\mathcal{R}_4$ switches on only when the agent changes the lane, as maintaining a safe distance while staying in the lane is contingent on the following vehicle. Specifically, if $\Delta p_{FS}^\prime < s^*_{FS}$ and $|a^{lc}| \neq 0$, then $\mathcal{R}_4 < 0$ is satisfied. 

Finally, the fifth reward component $\mathcal{R}_5$ is related to the accident (e.g., inter-vehicle crash, changing the nonexistent lanes) and is defined as follows.
\begin{align}
    \mathcal R_5
    &= 
    \begin{cases}
    0 & \mathrm{accident}~\mathrm{not}~\mathrm{happened}\\
    -1 & \mathrm{accident}~\mathrm{happened} 
    \label{eqn:R5}
    \end{cases}
\end{align}
The agent receives the penalty when the agent's action contributes to the accident (i.e., if the agent cannot continue driving). Typically, the fifth balancing weight $\eta_5$ is assigned the highest value among coefficients $\eta_1, \cdots, \eta_5$. 

Regardless of the driving scenario, the agent takes the decision-making process using the proposed POMDP. Note that the proposed POMDP is designed for safe and efficient driving without considering the routing of the destination.

\begin{figure*}[t]
    \centering
    \includegraphics[width=\textwidth]{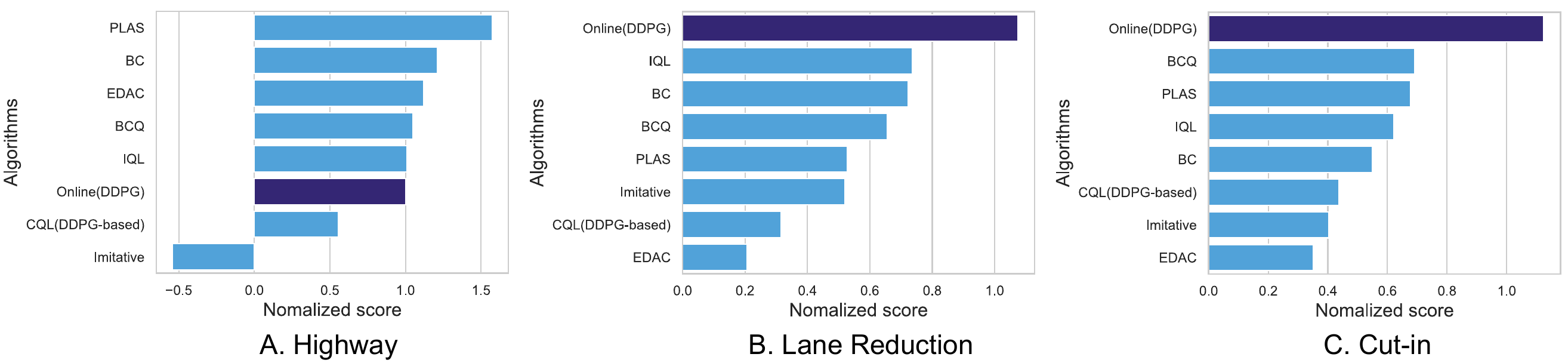}
    \caption{Normalized score for each algorithm per driving scenario. The average performance across synthetic datasets (except for \texttt{Human-like}) is provided to mitigate the impact of the dataset and to highlight the impact of the algorithm.}
    \label{fig:normalized}
\end{figure*}
\begin{figure*}[t]
    \centering
    \includegraphics[width=\textwidth]{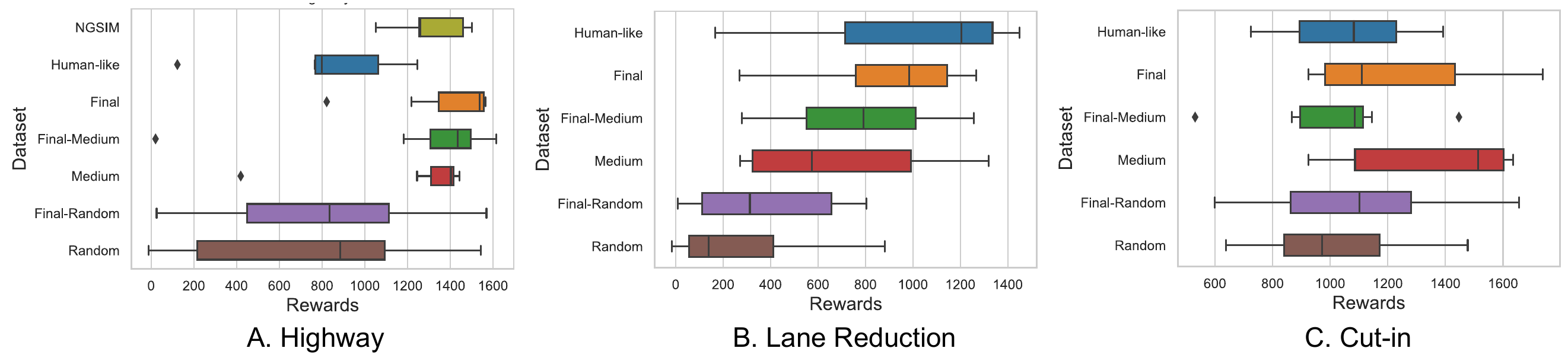}
    \caption{The IQR range of performance for each dataset per driving scenario. To mitigate the impact of the algorithms and to highlight the impact of the dataset, the IQR range includes the results of all algorithms. The diamonds indicate outliers, and the boxes represent the interquartile range. Herein, the start, middle, and end of the box indicate the quartile {1}, {2}, and {3}, respectively.
    }
    \label{fig:iqr}
\end{figure*}

\section{Benchmarking Baseline Performances}
This section provides the simulation results across all driving scenarios and datasets. 
To provide the performance baseline for autonomous driving scenarios, we performed extensive experiments and evaluated the performance of state-of-the-art offline reinforcement learning algorithms, including Behavioral Cloning (BC), imitative learning (DDPG + BC)~\cite{fujimoto2021minimalist}, batch constrained Q (BCQ)~\cite{fujimoto2019off}, conservative Q learning (CQL)~\cite{kumar2020conservative}, implicit Q learning (IQL)~\cite{kostrikovoffline}, ensemble-diversified actor-critic (EDAC)~\cite{an2021uncertainty}, and policy in the latent action space (PLAS)~\cite{zhou2021plas}.\footnote{Note that all algorithms cannot aim to work on hybrid action space but well work on discrete action space through simple quantization.} As discussed in section~\ref{sec:dataset}, we utilize the DDPG as the online reinforcement learning algorithm.\footnote{\textbf{Why not use the TD3?} We utilize the negative reward trick to prevent adverse effects on the approximation process when considering negative and positive reward signals. In such a setup, TD3~\cite{fujimoto2018addressing}, a higher version of DDPG, can lead to empirically overestimating the absolute Q-value. Therefore, we employ the DDPG.}

\subsection{Metrics} 
We use three evaluation metrics to verify the algorithm's and dataset's performance per driving scenario.

1) \textbf{Normalized Score}: 
It measures the effectiveness of offline reinforcement learning algorithms when the expert online agent's performance $\mathrm{score_{final}}$ is set as a baseline. The normalized score is calculated as $$\mathrm{normalized}~\mathrm{score} = \frac{(\mathrm{score} - \mathrm{score}_\mathrm{random})}{(\mathrm{score}_\mathrm{final} - \mathrm{score}_\mathrm{random})},$$ where $\mathrm{score_{random}}$ represents the score of a random policy. 

2) \textbf{Inter-quartile Range (IQR)}:
We use the IQR to display the performance range for each dataset~\cite{agarwal2021deep}. The IQR mitigates the impact of outliers, so it can be a more robust and statistically efficient measure than the median or mean.

\subsection{Simulation Results} The results offer the following advantages: 1) facilitating autonomous driving research by exploring offline reinforcement learning possibilities; 2) discussing the usability of the actual dataset in offline reinforcement learning.
All simulation results are presented in Table~\ref{tab:flow}. 
The presented score is the average normalized score over five random seeds. Each policy is evaluated by averaging the performance over ten executions.

\textbf{Performance over algorithms}:
Fig.~\ref{fig:normalized} shows the average normalized score per driving scenario and algorithm. This evaluation result presents the average performance across datasets. We comprehend that online reinforcement learning performance generally outperforms offline reinforcement learning performance. It is intuitive because online reinforcement learning can guarantee higher performance if there is enough exploration period.

\textbf{Performance over dataset}:
Fig.~\ref{fig:iqr} presents the IQR of the evaluation score per driving scenario and dataset. This evaluation results include the performance of all algorithms. In Fig.~\ref{fig:iqr}\textbf{B-C}, the results empirically imply two interesting insights: 1) the performance of {NGSIM} and {Human-like} datasets are comparable to synthetic datasets (close to {Final} and {Medium} for the most part), and 2) The performance ranks of the synthetic datasets is fair-minded ({Final} $>$ {Final-Medium} $>$ {Medium} $>$ {Final-Random} $>$ {Random}). On the other hand, In Fig.~\ref{fig:iqr}\textbf{C}, the average performance of the dataset is far from expected. The {Final} dataset contains samples with the highest performance, but the overall performance is highest in the {Medium}. 

\section{Conclusion}
This study presents an autonomous driving framework based on offline reinforcement learning, accompanied by benchmark performances and datasets that are readily accessible and reproducible. The driving scenarios within the FLOW framework have been expanded to include three realistic road structures: Cut-in, Lane Reduction, and Highway. A unified POMDP has been developed, applicable to all driving scenarios. The contribution of this work extends beyond providing synthetic datasets obtained from online reinforcement learning for each driving scenario, as it also includes pre-processed real-world driving datasets, such as the NGSIM dataset, aligned with the proposed POMDP. As a result, interesting insights have been obtained through the analysis of the results. 

In summary, the primary aim of this study is to utilize pre-collected datasets to facilitate research in the field of autonomous driving with offline reinforcement learning. 
We expect to accelerate progress in this domain and open up new avenues for further exploration.





\bibliographystyle{IEEEtran}
\bibliography{ref.bib}

\end{document}